\documentclass{article}
\usepackage{ijcai26}

\usepackage{times}
\usepackage{soul}
\usepackage{url}
\usepackage[hidelinks]{hyperref}
\usepackage[utf8]{inputenc}
\usepackage[small]{caption}
\usepackage{graphicx}
\usepackage{xcolor}
\usepackage{amsmath, amssymb, bm}
\usepackage{amsthm}
\usepackage{booktabs}
\usepackage{algorithm}
\usepackage{algorithmic}
\usepackage{subcaption}
\usepackage[switch]{lineno}

\title{Scaling Decision-Focused Learning to Large Problems \\ with Lagrangian Decomposition}
 \author{
 Stéphane Eilles-Chan Way$^{1,2}$
 \and
 Hugo Percot$^{1,2}$\and
 Quentin Cappart$^{1,3,4}$\and \\
 Tias Guns$^5$\and
 Louis-Martin Rousseau$^{1}$\\
 \affiliations
 $^1$Polytechnique Montréal, Montreal, Canada \\
 $^2$Ecole Polytechnique, Palaiseau, France \\
 $^3$UCLouvain, Louvain-la-Neuve, Belgium \\
 $^4$Mila - Québec AI Institute, Montreal, Canada\\
 $^5$KU Leuven, Leuven, Belgium\\
 \emails
 \{stephane.eilles-chan-way,hugo.percot\}@polytechnique.edu, \\
  tias.guns@kuleuven.be, \\
 \{quentin.cappart,louis-martin.rousseau\}@polymtl.ca
}

\begin{document}

\maketitle

\begin{abstract}
Decision-focused learning has shown great promise for addressing predict-then-optimize problems, particularly in the presence of under-specified models. However, its practical deployment is often hindered by high computational costs and limited scalability, as it requires solving a constrained optimization problem for each training instance at every iteration. To address these challenges, we propose a novel framework that incorporates Lagrangian decomposition into the decision-focused learning paradigm. Specifically, we introduce a new surrogate objective along with two loss functions for evaluating and training the underlying prediction model. We further propose two variants of our approach, which offer different trade-offs between computational efficiency and solution quality. Our framework can be seamlessly integrated with standard decision-focused learning methods, including \textit{Smart Predict-then-Optimize} (\texttt{SPO+}) and \textit{Implicit Maximum Likelihood Estimation} (\texttt{IMLE}).
Through experiments on two standard benchmarks, the multi-dimensional knapsack problem and quadratic portfolio optimization, we demonstrate that our approach achieves competitive performance while remaining amenable to parallelization. In particular, it consistently outperforms traditional decision-focused learning methods  on large-scale instances, involving up to eight times more variables than those typically considered in related work. The implementation is available at \url{https://github.com/corail-research/DFL-LD}.

%\keywords{Decision-focused learning  \and Lagrangian decomposition}

\end{abstract}

\section{Introduction}

Many real-world problems can be formulated as \textit{constrained optimization problems} (COPs), where the goal is to identify an optimal decision that maximizes or minimizes an objective function subject to a set of constraints. Both the objective function and the feasible region are typically parametric, depending on decision variables as well as contextual parameters. Although decades of research have produced highly efficient solvers for scenarios in which these contextual parameters are known, such information is often unavailable in practical applications.
Consequently, in most realistic settings, the contextual parameters must be predicted from external observations, for example, using a machine learning model, before being used by an optimization solver to determine the downstream optimal decision. This framework is commonly referred to as the \textit{predict-then-optimize} paradigm.
In such situations, suboptimal downstream decisions generally stem not from the limitations of the COP solver, but from inaccuracies in the predicted contextual parameters. Designing a predictive model that leads to near-optimal downstream decisions is therefore a central challenge. Traditional machine learning approaches, which optimize for predictive accuracy alone, often fall short because they ignore the combinatorial structure and decision sensitivity inherent to the underlying optimization problem.

To overcome this limitation, the \textit{decision-focused learning} (DFL) paradigm has emerged. Rather than training the predictive model to minimize a standard prediction loss, this approach explicitly incorporates the quality of the downstream decisions into the learning process, thereby aligning prediction and optimization objectives. 
Although Elmachtoub et al.~\cite{elmachtoub_estimate-then-optimize_2023} demonstrated that \textit{prediction-focused learning} (i.e. treating estimation and decision making as independent
steps) performs best when the underlying predictive model is well-specified, it is outperformed by DFL approaches in the presence of epistemic or aleatoric uncertainty (i.e. when the model is under-specified or the data are noisy).
Early work on DFL proposed direct differentiation through the \textit{argmin} operator, initially limited to unconstrained relaxations~\cite{gould_differentiating_2016}, and later extended to more general settings via differentiation of the Karush-Kuhn-Tucker conditions~\cite{amos_optnet_2017}.
Subsequent developments introduced several alternative formulations, including the \texttt{SPO+} surrogate loss \cite{elmachtoub_smart_2021} \texttt{IMLE} \cite{niepert_implicit_2021}, 
\texttt{CaVE}~\cite{CaVE2024}, \texttt{LAVA}~\cite{berden2025solverfreedecisionfocusedlearninglinear}, and the neuro-symbolic \texttt{E-PLL}~\cite{defresne2023scalablecouplingdeeplearning} for discrete graphical models, among others. For a comprehensive review and benchmark comparison, we refer the reader to the survey by Mandi et al. \cite{mandi_decision-focused_2024}. While DFL effectively aligns learning with decision quality, it suffers from substantial computational overhead: at each training step and for every instance, a constrained optimization problem must be solved. Since these problems are often NP-hard, and as the number of solver calls scales with both dataset size and the number of epochs, the computational cost increases sharply with instance size. This turns the training process into a sequence of expensive combinatorial optimizations that severely limit scalability and usability to solve large-scale instances.

Separately, \textit{Lagrangian decomposition}~\cite{guignard_lagrangean_1987} (LD) is a scalable optimization technique that has been successfully applied to large-scale combinatorial problems, in both mixed-integer programming and constraint programming contexts. Compared to the standard Lagrangian relaxation, LD can provide tighter dual bounds and handle arbitrary constraint structures, but at the expense of an increased computational cost: Guignard and Kim~\cite{guignard_lagrangean_1987}~(see Theorem 3.1) prove that for linearly-constrained problems the LD bound is at least as tight as the Lagrangian relaxation.
The key idea behind LD is to duplicate variables and partition the initial problem into independent subproblems. Ideally, these subproblems should preserve the combinatorial nature of the original formulation while being significantly faster to solve.
The solutions of the subproblems can then be combined to derive a valid dual bound on the original problem. However, this bound depends on a vector of Lagrangian multipliers that coordinate the subproblems; minimizing it amounts to optimizing these multipliers, typically through subgradient methods~\cite{shor2012minimization}. Lagrangian decomposition also integrates naturally with branch-and-bound algorithms: as noted by Guignard and Kim, disagreements between variable copies  directly suggest branching candidates, yielding a natural dichotomic branching rule. In the context of constraint programming, Hà et al.~\cite{ha_general_2015} demonstrated that LD can serve as an automatic, though computationally expensive, bounding mechanism that substantially reduces the size of the search tree. More recently, Bessa et al.~\cite{bessa2025learning} showed that machine learning techniques can be leveraged to warm-start the subgradient optimization, thereby improving the efficiency of the LD process. 

Inspired by this recent line of work, and by other contributions highlighting the benefits of combining learning with Lagrangian methods~\cite{parjadis2024learning,abbas2024doge}, we propose to integrate Lagrangian decomposition into a decision-focused learning  pipeline.
This integration enables DFL to scale effectively to very large combinatorial optimization problems. Our approach contrasts with most existing DFL research, which primarily focuses on developing new differentiation techniques. Instead, we introduce a general methodology that can be seamlessly combined with any existing differentiation algorithm, providing a flexible and scalable framework for decision-focused learning.

Our contributions are as follows: (1) a methodology to leverage Lagrangian decomposition into DFL, yielding a new surrogate function to optimize, (2) two loss functions to train a prediction model, and (3) the integration of our approach on two common DFL techniques (\texttt{SPO+} and \texttt{IMLE}).
The experiments are conducted on instances of the multi-dimensional knapsack problem with up to 300 variables and 10 constraints, which is roughly six times larger than what is commonly considered in prior work, and on instances of the portfolio optimization problem with up to 400 variables, representing an 8-fold increase compared to typical benchmarks. 
To our knowledge, this work constitutes the first attempt to incorporate Lagrangian decomposition into a DFL pipeline. 

\section{Problem Setting}

In many industrial applications, certain parameters of the constrained optimization problem are uncertain and must be inferred from contextual data, commonly referred to as \textit{features}. The settings we consider involve estimating these parameters through predictive inferences made by machine learning models, after which the final decisions are obtained by solving the corresponding optimization problems based on these predictions. In this framework, the overall decision-making process can be formulated as a \textit{predict-then-optimize problem}, where the optimization depends on parameters inferred from observed features.

 Let \(\mathbf{c} \in \mathbb{R}^{k}\) denote a vector of $k$ unknown parameters, 
 $\mathbf{X}$ decision variables defined over domain $D$, $C = (C_1,\dots,C_m)$ a set of $m$ constraints, and $f : D\times\mathbb{R}^k \to \mathbb{R}$ an objective function. 
 The problem we consider is as follows:
\begin{equation}
\max_{\mathbf{X} \in D} \; \left\{ f(\mathbf{X}, \mathbf{c}) \left\vert \; \bigwedge_{i=1}^{m} C_i(\mathbf{X}) \right. \right\}. \tag{$\mathcal{P}_c$}
\label{eq:COP-param}
\end{equation}
The optimal solution is denoted by $\mathbf{X}^*$. 
Although the optimal solution may not be unique, we only require a consistent tie-breaking rule so that $\mathbf{c} \mapsto \mathbf{X}^*(\mathbf{c})$ is well-defined as a function. In practice, this is satisfied since the solver returns a single deterministic solution.

In practice, the true cost parameters \(\mathbf{c}\) are not directly observable and must be inferred from feature vectors \(\mathbf{z}\in \mathbb{R}^q\). This inference is performed using a predictive model $\Omega_\theta : \mathbb{R}^q \to \mathbb{R}^k$ which produces an estimate $\mathbf{\hat{c}}=\Omega_{\theta}(\mathbf{z})$, where $\theta$ denotes the parameters of the model. The decision ultimately implemented is referred to as the \textit{prescriptive solution}, denoted \(\mathbf{X}^{*}(\hat{\mathbf{c}})\). This solution is generally suboptimal, and its quality can be assessed through the regret, defined as:
\begin{equation}
\mathcal{L}_{\text{regret}}(\mathbf{\hat{c}},\mathbf{c}) = f(\mathbf{X}^*(\mathbf{c}), \mathbf{c}) - f(\mathbf{X}^*(\mathbf{\hat{c}}), \mathbf{c}).
\label{eq:regret}
\end{equation}

Our objective is to learn a predictive model that produces prescriptive solutions as close as possible to the true optimal ones over a set of historical observations, or equivalently, to minimize the expected regret.
To perform this prediction task, we employ neural networks as a differentiable predictive models. Training is carried out using gradient descent, which requires a loss function  whose gradient can be efficiently computed or approximated. The choice of an appropriate loss is critical, as it has a major impact on the overall performance of the model.

\textit{Prediction-focused learning}  is arguably the most intuitive approach to address this problem. In this paradigm, estimation and decision making are treated as two fully independent steps. The predictive model $\Omega_{\theta}$ is trained to estimate $\mathbf{c}$ as accurately as possible by minimizing a prediction loss, typically the mean squared error between the true parameters 
$\mathbf{c}$ and their estimates $\mathbf{\hat{c}}$. Formally, this loss is defined as follows:
\begin{equation}
\label{eq:MSE}
\mathcal{L}_{\text{MSE}} (\mathbf{\hat{c}},\mathbf{c)} = \|\mathbf{c}-\mathbf{\hat{c}}\|^2.
\end{equation}
However, this approach has several limitations, particularly in under-specified settings, as demonstrated by Elmachtoub et al.~\cite{elmachtoub_estimate-then-optimize_2023}.
This is why decision-focused learning is often more appropriate, as it directly targets the quality of the prescriptive decision. In this setting, the regret, as defined in Equation~\eqref{eq:regret}, is used as the loss function. However, this formulation introduces new challenges.
First, computing the gradient of Equation~\eqref{eq:regret} is difficult, and sometimes even meaningless, as $\frac{\partial\mathbf{X}^*}{\partial\mathbf{\hat{c}}}$ can potentially be null almost everywhere. 
This occurs, for example, when the mapping
$\mathbf{c}\mapsto\mathbf{X}^*(\mathbf{c})$ is piecewise constant, as is common in combinatorial optimization. Fortunately, prior work has proposed ways to obtain exact gradients in special cases, or useful gradient approximations in more general scenarios, thereby enabling DFL in principle. Second, DFL suffers from substantial computational cost. Because it relies on the prescriptive decision during training, a new COP must be solved for every instance at every epoch. When the underlying problem is NP-hard, these repeated solver calls often become the primary bottleneck in the training pipeline. As a result, classical DFL methods do not scale well to large COPs, whose complexity grows exponentially.
Our contribution is an automatic mechanism based on Lagrangian decomposition to tackle this challenge.

\section{Lagrangian Decomposition for DFL}
The key idea of Lagrangian decomposition  is to  provide a dual bound to a given multi-constrained optimization problem by solving mono-constrained ones. 
Let us start from \eqref{eq:COP-param} defined earlier. Let $\mathbf{X}_1$ refer to the initial variables.
We introduce $m-1$ new variables by duplicating $\mathbf{X}_1$, obtaining a vector $\tilde{\mathbf{X}} =(\mathbf{X}_1,\dots,\mathbf{X}_m)$ where each $\mathbf{X}_i$ keeps the same  domain as $\mathbf{X}_1$. We first observe that, by adding equality constraints, the original COP \eqref{eq:COP-param} is equivalent to: 
\begin{equation}
   \max_{\mathbf{\tilde{X}} \in D} \left\{ f(\mathbf{X}_1, \mathbf{c}) \left\vert \left(\bigwedge_{i=1}^{m} C_i(\mathbf{X}_i)\right)\wedge \left(\bigwedge_{i=2}^{m} \mathbf{X}_i=\mathbf{X}_1\right)\right. \right\}
   \label{eq:COP-duplicate}
\end{equation}

The notation $C_i(X_i)$  means that constraint $C_i$ is applied to $X_i$. This does not require any structural assumption on the constraints since each $X_i$ is a full copy of the original variable vector $X$ (same dimension and domain). We then relax the new equality constraints by adding penalty terms into the objective function with multipliers $\mu =(\bm{\mu}_2,\dots, \bm{\mu}_m)\in \mathbb{R}^{|\mathbf{X}_1|\times (m-1)}$. By relaxing the equality constraints, we obtain a dual bound: 
\begin{align}
   B(\mu, \mathbf{c}) = \max_{\mathbf{\tilde{X}}\in D} \Biggl\{ &f(\mathbf{X}_1, \mathbf{c}) \;+ \nonumber\\ &\sum_{i=2}^m \bm{\mu}_i \cdot (\mathbf{X}_1 - \mathbf{X}_i) \bigg| \bigwedge_{i=1}^{m} C_i(\mathbf{X}_i) \Biggr\}.
   \label{eq:COP-LD1}
\end{align}
We note that this bound is parametrized by the multipliers $\mu$. Those are commonly referred to as \textit{Lagrangian multipliers}. Since each constraint $C_i$ acts on a distinct variable and the objective function is separable on the variables, the computation of $B(\mu, \mathbf{c})$ can be decomposed into $m$ subproblems featuring only a single constraint:
\begin{align}
    B(\mu, \mathbf{c}) &= \Phi(\mu, \mathbf{c}) + \sum_{i=2}^m \Psi_i(\bm{\mu}_i), ~\text{where}\label{eq:LD_X1}\\
    \Phi(\mu, \mathbf{c}) &= \max_{\mathbf{X}_1 \in D} \left\{ f(\mathbf{X}_1, \mathbf{c}) + \sum_{i=2}^m \bm{\mu}_i \cdot \mathbf{X}_1 \left\vert \, C_1(\mathbf{X}_1)\right. \right\} \label{eq:LD_X2}\\
    \Psi_i(\bm{\mu}_i) &= \max_{\mathbf{X}_i \in D} \; \left\{-\bm{\mu}_i \cdot \mathbf{X}_i\left\vert \; C_i(\mathbf{X}_i)\right. \right\}, \forall i \in [\![2,m]\!].  \label{eq:LD_X3}
\end{align}

In practice, we are interested in finding the tightest possible dual bound, which requires computing the best multipliers i.e. $\mu^*(\mathbf{c})= \text{argmin}_\mu(B(\mu,\mathbf{c}))$. This can be done with a subgradient descent method to optimize the argmin, such as Shor's algorithm~\cite{shor2012minimization}. 
Let  $\tilde{\mathbf{X}}^*(\mu,\mathbf{c})=(\mathbf{X}_1^*(\mu,\mathbf{c}),\dots,\mathbf{X}_m^*(\mu,\mathbf{c}))$ be the solution of Equation~\eqref{eq:COP-LD1} for a given $\mu$.
A standard subgradient expression of $B(\mu,\mathbf{c})$ with regard to $\bm{\mu}_i$ is $(\mathbf{X}_1^*(\mu,\mathbf{c}) - \mathbf{X}_i^*(\mu,\mathbf{c}))$ for all ${i \in [\![2,m]\!]}$  \cite{bessa2025learning}.

\subsection{Defining Appropriate Loss Functions}
\label{subsec:LD_DFL}

Lagrangian decomposition provides a fast way to compute an upper bound, although the bound’s tightness depends on the chosen multipliers. Ideally, we would replace the COP-solving phase in the traditional DFL pipeline with Lagrangian decomposition using optimal multipliers, which yield the tightest dual bound.
Taking a step back, we note that Equation~(\ref{eq:COP-LD1}), when parameterized and evaluated with optimal multipliers, becomes:
\begin{align}
   B\big(\mu^*(\mathbf{c}),\mathbf{c}\big) =& \max_{\mathbf{\tilde{X}} \in D}\; \Biggl\{ f(\mathbf{X}_1,\mathbf{c}) + \nonumber\\&\sum_{i=2}^m \bm{\mu}_i(\bm{c}) \cdot (\mathbf{X}_1 - \mathbf{X}_i) \left\vert \; \bigwedge_{i=1}^{m} C_i(\mathbf{X}_i)\right. \Biggr\}.
   \label{eq:COP-LD2}
\end{align}
 This expression is a new COP parameterized by $\left(\mu^*(\mathbf{c}), \mathbf{c}\right)$. Therefore, we can reuse the tools from DFL, except that we need to handle the multipliers $\mu^*(\mathbf{c})$ as well, which act as parameters in Equation~\eqref{eq:COP-LD2} along $\mathbf{c}$, but are not given by the predictive model upstream.
 In contrast, they are obtained by a third-party optimization algorithm. 
The next step is to design an appropriate DFL-based loss function. Let $\mathbf{\tilde{X}}^*(\mu^*,\mathbf{c})=(\mathbf{X}_1^*(\mu^*,\mathbf{c}),\dots,\mathbf{X}_m^*(\mu^*,\mathbf{c}))$ denote the solution to Equation~(\ref{eq:COP-LD2}) and let 
$b : (\mathbf{\tilde{X}}, \mu, \mathbf{c}) \mapsto f(\mathbf{X}_1,\mathbf{c}) + \sum_{i=2}^m \bm{\mu}_i \cdot (\mathbf{X}_1 - \mathbf{X}_i)$ be a function mapping for Lagrangian decomposition. 
We propose two loss functions that integrate Lagrangian decomposition into a DFL pipeline:
\begin{align}
\mathcal{L}_1\left(\mathbf{\hat{c}},\mathbf{c}\right) = b(\mathbf{\tilde{X}}^*(\mu^*(\mathbf{\hat{c}}),&\mathbf{\hat{c}}), \mu^*(\mathbf{c}),\mathbf{c})\nonumber\\ &- b(\mathbf{\tilde{X}}^*(\mu^*(\mathbf{c}),\mathbf{c}),\mu^*(\mathbf{c}),\mathbf{c}) \label{eq:L1-orgin}
\end{align}
\begin{equation}
\mathcal{L}_2\left(\mathbf{\hat{c}},\mathbf{c}\right) = f(\mathbf{X}_1^*(\mu^*(\mathbf{\hat{c}}),\mathbf{\hat{c}}),\mathbf{c}) - f(\mathbf{X}_1^*(\mu^*(\mathbf{c}),\mathbf{c}),\mathbf{c})
    \label{eq:L2-orgin}
\end{equation}

\begin{figure*}
    \centering
    \includegraphics[width=0.8\linewidth]{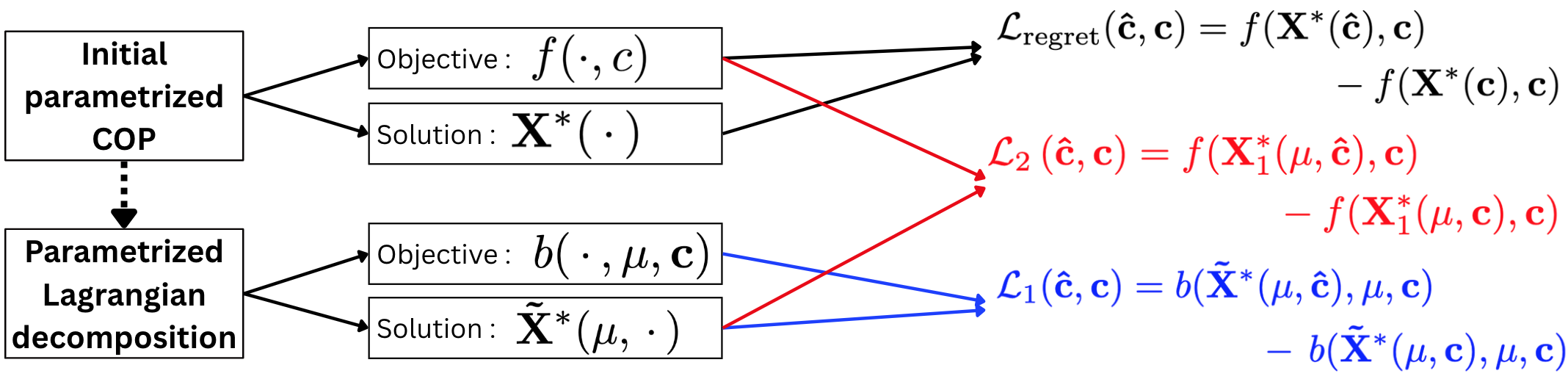}
    \caption{Schematic view of our method. The initial parameterized COP and the parameterized LD define two optimization problems, each with its own objective function and corresponding optimal solution. Each approach  selects a particular combination of these components to construct a loss function. The regret  $\mathcal{L}_{\text{regret}}$ denotes the classical DFL loss, while $\color{blue}\mathcal{L}_{1}$ and  $\color{red}\mathcal{L}_{2}$ are the new losses we propose.}
    \label{fig:placeholder}
\end{figure*}

The first loss ($\mathcal{L}_1$)
corresponds exactly to the regret associated with Equation~\eqref{eq:COP-LD2}. It involves the objective function $b(\mathbf{\tilde{X}}, \mu, \mathbf{c})$ where the Lagrangian multipliers appear explicitly. 
In contrast, in the second loss ($\mathcal{L}_2$), explicit penalty terms with the multipliers disappear since $f(\mathbf{X}, \mathbf{c})$ is used instead of $b(\mathbf{\tilde{X}}, \mu, \mathbf{c})$. However, the multipliers are still involved in the resolution of $\mathbf{X}_1^*$, so they are still ultimately considered.
Intuitively, $\mathcal{L}_2$ mirrors the regret defined from \eqref{eq:COP-param}, but it is evaluated at $\mathbf{X}_1^*(\mu^*,\mathbf{\hat{c}})$ rather than $\mathbf{X}^*(\mathbf{\hat{c}})$.  We use $\mathbf{X}_1^*$ (instead of the duplicated variables $\mathbf{\tilde{X}}^*$) because these variables are obtained from Equation~\eqref{eq:LD_X1}, which directly incorporates both the objective $f$ and the multipliers.  Since the bottleneck in DFL is the solving time of the COP at each iteration, we expect the two losses based on Lagrangian decomposition to be significantly faster. As a counterpart, as they are based on a relaxed problem (Eq.~\eqref{eq:COP-LD2}), the accuracy of the estimator $\mathbf{\mathbf{\hat{c}}}$ on the primal \eqref{eq:COP-param} is not guaranteed. The choice to use $f(\mathbf{X}, \mathbf{c})$ in $\mathcal{L}_2$ is motivated by the idea that it could strengthen the performance during evaluation on \eqref{eq:COP-param}.
Figure~\ref{fig:placeholder} illustrates how these two losses differ from the standard regret loss.

\subsection{Dealing with  Multipliers under Uncertainty}

Unfortunately, minimizing these two loss functions is challenging, mainly because the multipliers depend on the uncertain parameters, 
i.e., $\mu^*(\mathbf{\hat{c}})$. A first issue is that computing $\mu^*(\mathbf{\hat{c}})$ for each instance would significantly increase computational cost. Moreover, differentiating the losses (i.e., $\frac{\partial\mathcal{L}}{\partial\mathbf{\hat{c}}}$)  requires computing both $\frac{\partial \mathbf{X}_1^*}{\partial\mathbf{\hat{c}}}$ and $\frac{\partial\mu^*}{\partial\mathbf{\hat{c}}}$. As a result, optimizing these losses becomes a bilevel optimization problem. 
We address this challenge by \textit{fixing} the Lagrangian multipliers rather than optimizing over $\mu^*(\mathbf{\hat{c}})$. 
With fixed multipliers, the optimal solutions to the subproblems $(\mathbf{X}^*_2,\dots,\mathbf{X}^*_m)$  no longer depend on $\mathbf{\hat{c}}$ (Eq.~\eqref{eq:LD_X3}).
The next step is to decide how to set these multipliers. A natural choice is to fix them to their optimal values under the true cost parameters, i.e., we set $\mu^*(\mathbf{\hat{c}})= \mu^*(\mathbf{c})$. Intuitively, this choice reflects the multipliers we would like to obtain at the end of training. The two corresponding losses  can then be reformulated as follows:
\begin{align}
\mathcal{L}^\text{new}_1\left(\mathbf{\hat{c}},\mathbf{c}\right) &= \sigma(\mathbf{X_1^*}(\mu^*(\mathbf{c}),\mathbf{\hat{c}}),\mu^*(\mathbf{c}),\mathbf{c})\notag\\
    &-\sigma(\mathbf{X_1^*}(\mu^*(\mathbf{c}),\mathbf{c}),\mu^*(\mathbf{c}),\mathbf{c}) \label{eq:l1new} \\
    \mathcal{L}^\text{new}_2\left(\mathbf{\hat{c}},\mathbf{c}\right) &= f(\mathbf{X_1^*}(\mu^*(\mathbf{c}),\mathbf{\hat{c}}),\mathbf{c})\notag\\
    &-f(\mathbf{X_1^*}(\mu^*(\mathbf{c}),\mathbf{c}),\mathbf{c}) \label{eq:l2new} 
\end{align}
where $\sigma : (\mathbf{X_1}, \mu, \mathbf{c}) \mapsto f(\mathbf{X_1},\mathbf{c}) + \sum_{i=2}^m \bm{\mu}_i \cdot \mathbf{X_1}$. The resulting training algorithm, referred to as $\texttt{trainingSingle}$ since it relies on a single fixed Lagrangian decomposition, is depicted in Algorithm~\ref{alg:DFL}. It illustrates how the predictive model $\Omega_\theta$ is trained using a dataset $\mathcal{S} = \{(\mathbf{z}_i, \mathbf{c}_i, \mathbf{X_1^*}(\mathbf{c}_i),\mu^*(\mathbf{c}_i)\}_{i=1}^n$. Recall that $\mathbf{z}_i$
denotes the feature vector of data sample $i$. We note that constructing this training set requires computing the Lagrangian multipliers, via a subgradient method, for all instances in the training set and for multiple cost parameter values, which can be computationally expensive. Instead, we propose to terminate the subgradient procedure after a predefined number of iterations and to use the resulting potentially suboptimal multipliers. As we will show in Section~\ref{sec:comp_exp}, this early termination still leads to competitive results in practice.
The model weights are initialized randomly (Line \ref{eq1:l1}), and training proceeds for a fixed number of epochs. Each data sample is processed once per epoch. For every sample, we first predict the cost parameters from the feature vector (Line \ref{eq1:l3}), then execute the Lagrangian decomposition of the problem (Line \ref{eq1:l4}), and finally solve the resulting subproblems (Line \ref{eq1:l5}). As explained previously, only the variables $\boldsymbol{X_1}^*$ are required to compute the loss. The weights are then updated by backpropagating the loss using the chain rule (Line \ref{eq1:l6}). 
The term $\frac{\partial \mathcal{L}}{\partial \boldsymbol{X_1}^*}$ is obtained directly from Eq.~\eqref{eq:l1new} or Eq.~\eqref{eq:l2new} and $\frac{\partial \hat{\boldsymbol{c}_i}}{\partial \theta}$ is computed by differentiating through $\Omega_\theta$, e.g., a neural network. 
Concerning the term $\frac{\partial \boldsymbol{X_1}^*(\hat{\boldsymbol{c}}_i,\mu)}{\partial \hat{\boldsymbol{c}}_i}$, we emphasize that the main advantage of our approach is that we do not need anymore to consider all the variables as in standard DFL. The LD subproblem is easier to solve and differentiate because it is smaller and has a simpler structure. This gradient can be obtained using classical DFL differentiation techniques or, when the subproblem has exploitable structure, via dedicated solver-free differentiation methods. These two situations are later considered in our experiments. Since our approach can be combined with any differentiation technique, it provides  flexibility in selecting the most suitable one for a given use case. Note that some  DFL differentiation techniques, e.g. \texttt{SPO+}, provide the straightforward gradient $\frac{\partial \mathcal{L}}{\partial \hat{\boldsymbol{c}}_i}$ instead of $\frac{\partial \boldsymbol{X_1}^*(\hat{\boldsymbol{c}}_i,\mu)}{\partial \hat{\boldsymbol{c}}_i}$, which is still compatible with Algorithm~\ref{alg:DFL}.

\begin{algorithm}[tb]
\caption{$\texttt{trainingSingle}(\mathcal{S},\Omega_\theta,\mathcal{P},\mathcal{L},\alpha,E)$}
\label{alg:DFL}
\begin{algorithmic}[1]
\REQUIRE Dataset $\mathcal{S} = \{(\mathbf{z}_i, \mathbf{c}_i, \mathbf{X_1^*}(\mathbf{c}_i),\mu^*(\mathbf{c}_i))\}_{i=1}^n$.
\REQUIRE Predictive model $\Omega_\theta$.
\REQUIRE Combinatorial problem $\mathcal{P} = \langle \mathbf{X},D,C,f \rangle$.
\REQUIRE Loss function $\mathcal{L} \in \{\mathcal{L}_1^\text{new},\mathcal{L}^\text{new}_2\}$.
\REQUIRE learning rate $\alpha$, number of training epochs $E$.
\STATE $\theta$ := $\texttt{randomInitialization}() \label{eq1:l1}$
    \FOR{\textbf{each} epoch  \textbf{from} $1$  \textbf{to} $E$}
        \FOR{\textbf{each} sample $i \in \mathcal{S}$}
            \STATE $\mu := \mu^*(\mathbf{c}_i)$ \label{eq1:l2}
            \STATE $\hat{\boldsymbol{c}_i}$  := $\Omega_\theta(\boldsymbol{z}_i)$ \label{eq1:l3}
            \STATE $\Phi$ := $\texttt{LD}(\mathcal{P})$ \COMMENT{Eq.~\eqref{eq:LD_X2}} \label{eq1:l4}
            \STATE $\boldsymbol{X_1}^*$ := $\texttt{solve}(\Phi, \mu, \hat{\boldsymbol{c}_i})$ \COMMENT{Subproblem solving} \label{eq1:l5}
            \STATE $\theta$ := $\theta - \alpha \big( \frac{\partial \mathcal{L}}{\partial \boldsymbol{X_1}^*} \cdot  
            \frac{\partial \boldsymbol{X_1}^*(\hat{\boldsymbol{c}}_i,\mu)}{\partial \hat{\boldsymbol{c}}_i}\cdot  
            \frac{\partial \hat{\boldsymbol{c}_i}}{\partial \theta} \big)$ \COMMENT{Chain rule} \label{eq1:l6}
        \ENDFOR
    \ENDFOR
\RETURN $\Omega_\theta$
\end{algorithmic}
\end{algorithm}

\subsection{Handling Multiple Decompositions}

Our current approach considers only the subset $\boldsymbol{X}_1 \in \tilde{\mathbf{X}}$ when computing the loss. 
Although this choice improves overall computational efficiency, it raises an important question: the selection of variables to serve as the basis for duplication (see Eq.~\eqref{eq:COP-duplicate}) is inherently arbitrary. This nontrivial design choice can have a significant impact on the performance of our method. We propose to mitigate this issue by dynamically changing, during training, which variables $\boldsymbol{X}_i \in \tilde{\mathbf{X}}$ are used for duplication and by leveraging multiple decompositions.
A problem with $m$ constraints then admits $m$ possible decompositions.
Let us denote by $\boldsymbol{X}_1^{(d)}$ and $\boldsymbol{X}_1^{*(d)}$ the variables and their optimal values in the corresponding subproblem for decomposition $d \in [\![1,m]\!]$, respectively. Their associated losses $ \mathcal{L}_1^{(d)}$, and $\mathcal{L}_2^{(d)}$ are defined as follows.
\begin{align}
\mathcal{L}_1^{(d)}\left(\mathbf{\hat{c}},\mathbf{c}\right) =& \sigma(\mathbf{X_1^*}^{(d)}(\mu^*(\mathbf{c}),\mathbf{\hat{c}}),\mu^*(\mathbf{c}),\mathbf{c})\notag\\
&-\sigma(\mathbf{X_1^*}^{(d)}(\mu^*(\mathbf{c}),\mathbf{c}),\mu^*(\mathbf{c}),\mathbf{c})\\
\mathcal{L}_2^{(d)}\left(\mathbf{\hat{c}},\mathbf{c}\right) =& f(\mathbf{X_1^*}^{(d)}(\mu^*(\mathbf{c}),\mathbf{\hat{c}}),\mathbf{c}) \notag\\&-f(\mathbf{X_1^*}^{(d)}(\mu^*(\mathbf{c}),\mathbf{c}),\mathbf{c}).
\end{align}

Intuitively, these updated losses convey more information about the underlying problem. We propose to select which decomposition to use at random at each training iteration. 
As a drawback, since each decomposition $d$ selects a different variable and constraint for the main subproblem, it yields different Lagrangian multipliers $\mu^{*(d)}(\mathbf{c}_i)$ and subproblem solutions $\mathbf{X_1^*}^{(d)}(\mathbf{c}_i)$. We therefore need to construct a separate training set for each decomposition,
$\mathcal{S}^{(d)} = \{(\mathbf{z}_i, \mathbf{c}_i, \mathbf{X_1^*}^{(d)}(\mathbf{c}_i),\mu^{*(d)}(\mathbf{c}_i))\}_{i=1}^n$,
which is more time-consuming, but can still be performed offline. Moreover, these $m$ constructions are fully independent and thus easily parallelizable. The resulting training algorithm is depicted in Algorithm~\ref{alg:DFL_multiple}. It simply consists in
executing Algorithm~\ref{alg:DFL} (without the random initialization of the model parameters $\theta$) with a randomly selected decomposition at each epoch (Line \ref{eq2:l5}).

\begin{algorithm}[!ht]
\caption{$\texttt{trainingMultiple}()$}
\label{alg:DFL_multiple}
\begin{algorithmic}[1]

\REQUIRE Number of decompositions $m$.
\REQUIRE $\mathcal{S}^{(d)} = \{(\mathbf{z}_i, \mathbf{c}_i, \mathbf{X_1^*}(\mathbf{c}_i)^{(d)},\mu^*(\mathbf{c}_i)^{(d)})\}_{i=1}^n$.
\REQUIRE Predictive model $\Omega_\theta$.
\REQUIRE Combinatorial problem $\mathcal{P} = \langle \mathbf{X},D,C,f \rangle$.
\REQUIRE Loss function $\mathcal{L}^{(d)} \in \{\mathcal{L}^{(d)}_1,\mathcal{L}^{(d)}_2\}~\forall d \in [\![1,m]\!]$.
\REQUIRE learning rate $\alpha$, number of training epochs $E$.
\STATE $\theta$ := $\texttt{randomInitialization}() \label{eq2:l1}$
    \FOR{\textbf{each} epoch  \textbf{from} $1$  \textbf{to} $E$}
        \STATE $d := \texttt{randomSelection}(1,m)$ \label{eq2:l3}
        \STATE $\Omega_\theta := \texttt{trainingSingle}(S^{(d)},\Omega_\theta,\mathcal{P},\mathcal{L}^{(d)},\alpha,1)$ \label{eq2:l5}
    \ENDFOR
\RETURN $\Omega_\theta$

\end{algorithmic}
\end{algorithm}

\section{Case Studies}

This paper addresses two challenging problems in the DFL literature: the multi-dimensional knapsack problem and the quadratic portfolio optimization problem. These benchmarks are chosen to enable direct comparison with the work of Elmachtoub and Grigas~\cite{elmachtoub_smart_2021} and Mandi et al.~\cite{mandi2020smart}.

\paragraph{Multi-dimensional knapsack.}
It is an extension of the classic knapsack problem. In the multidimensional knapsack problem, we are given $m$ constraints and $n$ items, each associated with a profit $\mathbf{c}\in\mathbb{R}^{n}$ and multiple weights $(\mathbf{w}^1,\dots, \mathbf{w}^m) \in\mathbb{R}^{n\times m}$. Each constraint $k \in \{1,\dots,m\}$ represents a capacity limit $b_k$ for a different dimension. The objective is to select a subset of items that maximizes total profit while ensuring that the total weight in each dimension does not exceed the corresponding capacity constraint.
All benchmark instances follow the protocol explained by~\cite{tang2023pyepopytorchbasedendtoendpredictthenoptimize} and summarized in the survey of Mandi et al.~\cite{mandi_decision-focused_2024}. For each data point $i$, shared features~$\mathbf{z}_i\in\mathbb{R}^{p}$ are generated and then transformed into item profits $(c_{i,j})_{j=1,\dots,n}$ via the following  projection:
\begin{equation}
\label{eq:gen_knapsack}
c_{i,j} =\Bigl\lceil
  \bigl[\tfrac{5}{3.5^{t}}\,
        (\tfrac{1}{\sqrt{p}}(\mathcal{B}\mathbf{z}_{i})_j+3)^{t}+1\bigr]
  \,\epsilon_{i,j}
\Bigr\rceil.
\end{equation}
Here, $\epsilon_{i,j}$ is sampled uniformly from the interval $[1-\bar{\epsilon},\,1+\bar{\epsilon}]$, and $\mathcal{B}$ denotes a random projection matrix. Value $t$ is used to parametrize the degree of the projection. Item weights are sampled uniformly from $3$ to $8$
and remain fixed across the dataset, ensuring that uncertainty affects only the objective coefficients. The capacity limits $b_k$ equal the half of the sum of each constraint weights $(w^k_1,\dots, w^k_n)$ and also remain fixed.
We use $m = 10$,  $n \in \{100, 200, 300\}$, $t = 8$, $p = 12$ and $\epsilon = 0.5$,
corresponding to substantially larger problem instances than those typically considered in the literature. For example, the recent survey by Mandi et al. reports knapsack instances with only a single constraint and at most 45 items. The predictive model $\Omega_\theta$ is a simple linear regression. The sub-problems are one-dimensional knapsack problems and are solved using \texttt{Gurobi}~\cite{gurobi} (version 12.0.0). Finally, for each size $n$, 1300 instances are generated and split as follows: 200 for training, 100 for validation and 1000 for the final evaluation.

\paragraph{Quadratic portfolio optimization.}

Given predicted asset returns $\mathbf{c}\in\mathbb{R}^{n}$ and a positive semidefinite covariance matrix  $\Sigma\in\mathbb{R}^{n\times n}$ with average entry $\bar{\Sigma}$, the decision maker chooses portfolio weights $\mathbf{x}\in\mathbb{R}^{n}$ to maximize expected return, $\mathbf{c}^{\top}\mathbf{x}$, while respecting a risk budget, $\mathbf{x}^{\top}\Sigma\mathbf{x} \le \gamma \bar{\Sigma}$, and fully investing the available capital, $\mathbf{1}^{\top}\mathbf{x} = 1$. Again, the instances are generated following the procedure detailed by 
 Mandi et al.  using the hyperparameters they recommend.  In particular, five features and a simple linear regression for the predictive model are used. We consider instances with $n = \{50, 200, 400\}$ assets. For comparison, \cite{elmachtoub_smart_2021} consider instances with at most 50 assets. The sub-problems are solved using \texttt{Gurobi}~\cite{gurobi} (version 12.0.0). Finally, 10125 problems are generated and split as follows: 100 for training, 25 for validation and 10000 
 for the final evaluation. 
 Interestingly, when choosing the risk constraint as the main constraint, the resulting LD main subproblem:
\begin{equation}
\label{eq:portfolio_main}
\max_{\mathbf{X}_{1}\ge 0}
      \bigl(\mathbf{c}+\boldsymbol{\mu_2}\bigr)^{\top}\mathbf{X}_{1}
\quad
\text{s.t.}\quad
\mathbf{X}_{1}^{\top}\Sigma\mathbf{X}_{1}\le\gamma\bar{\Sigma}
\end{equation}
admits a closed-form optimal solution once we relax the nonnegativity constraint on the decision variable, given by
\begin{equation}
\label{eq:X1_exact}
\resizebox{.85\linewidth}{!}{$
            \displaystyle
            \mathbf{\bar{X}_1}^*(\mathbf{c}, \mu) = \sqrt{ \frac{\gamma \bar{\Sigma}}{\bigl(\mathbf{c}+\boldsymbol{\mu_2}\bigr)^\top \Sigma^{-1} \bigl(\mathbf{c}+\boldsymbol{\mu_2}\bigr)} }\, \Sigma^{-1} \bigl(\mathbf{c}+\boldsymbol{\mu_2}\bigr)
        $}.
\end{equation}
We refer the readers to Appendix~\ref{app:Exact} for the derivation of this exact closed-form optimal solution.
This allows us to design an exact differentiation scheme to compute this solution.
During training, we replace the solver-provided solution $\mathbf{X}_1^{*}(\mathbf{c},\mu)$ with the analytical expression $\mathbf{\bar{X}}_1^{*}(\mathbf{c},\mu)$ (Eq.~\eqref{eq:X1_exact}).  This eliminates the need for a solver and a generic DFL technique because its gradient can be obtained directly via automatic differentiation tools.

\begin{table*}[ht]
\centering
\setlength{\tabcolsep}{4pt}
\footnotesize
\begin{tabular}{l cc cc cc}
\toprule
\multicolumn{7}{c}{\textbf{Multi-dimensional knapsack} (10 constraints)} \\
\midrule
& \multicolumn{2}{c}{100 items} & \multicolumn{2}{c}{200 items} & \multicolumn{2}{c}{300 items} \\
\cmidrule(lr){2-3} \cmidrule(lr){4-5} \cmidrule(lr){6-7}
Approaches & Regret & Time (s) & Regret & Time (s) & Regret & Time (s) \\
\midrule
\texttt{MSE} & $7.796 \pm 0.034$ & $270\pm 44$ & $5.518 \pm 0.026$ & $383 \pm 37$ & $6.774 \pm 0.025$ & $387 \pm 20$ \\
\midrule
$\texttt{CaVE}$
& $3.680 \pm 0.052$ & $68 \pm 15$ & $3.617 \pm 0.034$ & $107 \pm 20$ & $3.466 \pm 0.038$ & $182 \pm 1$ \\
$\texttt{CaVE+}$
& $3.686\pm0.029$& $1177\pm816$& $3.553\pm0.020$ & $874\pm783$& $3.265\pm 0.015$ & $217\pm188$\\
$\texttt{LAVA}$
& $4.509 \pm 0.04$ & $4346 \pm 1053$& $4.552\pm0.02 $& $6198\pm 377$ & $4.558\pm0.05$ & $5878\pm683$ \\
\midrule
\texttt{SPO+} & $3.049 \pm 0.032$ & $4692 \pm 1128$ & $2.975 \pm 0.080$ & $4350 \pm 1724$ & $3.072 \pm 0.019$ & $1904 \pm 785$ \\
$\texttt{LD}(\texttt{SPO+},\texttt{static},\mathcal{L}_1,\infty)$
& $2.887 \pm 0.024$ & $ 41 \pm 2$ & $2.562 \pm               0.016$ & $42 \pm 4$ & $2.628\pm0.035$ & $107 \pm34$ \\
$\texttt{LD}(\texttt{SPO+},\texttt{multiple},\mathcal{L}_1,\infty)$
& \underline{$2.621\pm0.030$} & $8\pm1$ & \underline{$2.496 \pm 0.012$} & $1825\pm 571$ & $\mathbf{2.179\pm0.007}$ & $2485\pm963$ \\
\midrule
\texttt{IMLE} & $4.234 \pm 0.075$ & $5836 \pm 728$ & $3.285 \pm 0.047$ & $6489 \pm 327$ & $3.933 \pm 0.045$ & $5988 \pm 524$ \\
$\texttt{LD}(\texttt{IMLE},\texttt{static},\mathcal{L}_1,\infty)$
& $3.401 \pm 0.062$ & $3715\pm929$ & $3.309 \pm 0.027$ & $1087\pm158$ & $3.329 \pm 0.032$ & $4504\pm579$ \\
$\texttt{LD}(\texttt{IMLE},\texttt{static},\mathcal{L}_2,\infty)$
& $3.809\pm0.071$ & $5339\pm1210$ & $3.438 \pm 0.025$ & $1148\pm1289$ & $3.712 \pm 0.051$ & $64\pm12$ \\
$\texttt{LD}(\texttt{IMLE},\texttt{multiple},\mathcal{L}_1,\infty)$
& $2.807 \pm 0.019$ & $4813\pm964$ & $2.801 \pm 0.038$ & $610\pm464$ & $2.382 \pm 0.015$ & $6049\pm312$ \\
$\texttt{LD}(\texttt{IMLE},\texttt{multiple},\mathcal{L}_2,\infty)$
& $\mathbf{2.376 \pm 0.017}$ & $5293\pm614$ & $\mathbf{2.189 \pm 0.011}$ & $5905\pm352$ & \underline{$2.235 \pm 0.011$} & $6401\pm149$ \\

\midrule\midrule
\multicolumn{7}{c}{\textbf{Quadratic portfolio optimization} (2 constraints)} \\
\midrule
& \multicolumn{2}{c}{50 assets} & \multicolumn{2}{c}{200 assets} & \multicolumn{2}{c}{400 assets} \\
\cmidrule(lr){2-3} \cmidrule(lr){4-5} \cmidrule(lr){6-7}
Approaches & Regret & Time (s) & Regret & Time (s) & Regret & Time (s) \\
\midrule
\texttt{MSE} & $2.325 \pm 0.016$ & $12 \pm 3$ & $2.070 \pm 0.010$ & $993 \pm 765$ & $2.324 \pm 0.001$ & $4648 \pm 2827$ \\
\midrule
\texttt{SPO+} & $2.095 \pm 0.025$ & $58 \pm 7$ & $2.092 \pm 0.006$ & $4408 \pm 288$ & $2.087 \pm 0.010$ & $10277 \pm 699$ \\
$\texttt{LD}(\texttt{SPO+},\texttt{static},\mathcal{L}_1,\infty)$
& $2.122 \pm 0.021$ & $101 \pm 29$ & $14.833 \pm 0.250$ & $6685 \pm 124$ & $16.84 \pm 0.145$ & $11698 \pm 193$ \\
\midrule
\texttt{IMLE} & $2.064 \pm 0.008$ & $1450 \pm 1115$ & \underline{$2.032 \pm 0.009$} & $5586 \pm 907$ & \underline{$2.026 \pm 0.004$} & $11327 \pm 1252$ \\
$\texttt{LD}(\texttt{IMLE},\texttt{static},\mathcal{L}_1,\infty)$
& $2.285 \pm 0.011$ & $67 \pm 8$ & $11.028 \pm 0.143$ & $6339 \pm 291$ & $12.31 \pm 0.081$ & $11469 \pm 875$ \\
$\texttt{LD}(\texttt{IMLE},\texttt{static},\mathcal{L}_2,\infty)$
& $\mathbf{2.026 \pm 0.004}$ & $282 \pm 54$ & $2.101 \pm 0.020$ & $6465 \pm 477$ & $2.114 \pm 0.010$ & $11640 \pm 761$ \\
\midrule
$\texttt{LD}(\texttt{Exact},\texttt{static},\mathcal{L}_1,\infty)$
& $2.159 \pm 0.056$ & $712 \pm 853$ & $2.075 \pm 0.011$ & $69 \pm 49$ & $2.360 \pm 0.0191$ & $1975 \pm 1086$ \\
$\texttt{LD}(\texttt{Exact},\texttt{static},\mathcal{L}_2,\infty)$
& \underline{$2.040 \pm 0.009$} & $14 \pm 5$ & $\mathbf{1.969 \pm 0.010}$ & $1309 \pm 1683$ & $\mathbf{1.992 \pm 0.001}$ & $3614 \pm 3544$ \\
\bottomrule
\end{tabular}

\caption{Test-set relative regret and time-to-best validation model for multi-dimensional knapsack and quadratic portfolio optimization (mean across seeds $\pm 95\%$ CI). Relative regret values must be  multiplied by $10^{-2}$. Best results  are  \textbf{in bold}. Second best results are \underline{underlined}.}
\label{tab:bookmark-log}

\end{table*}

\section{Computational Experiments}
\label{sec:comp_exp}
This section presents the results of our computational experiments. All tests were conducted on a cluster of CPU nodes equipped with AMD EPYC 9654 and Intel Xeon Gold 6148 (Skylake) processors, with each run restricted to a single core and 4 GB of RAM. The optimization problems are solved using \texttt{Gurobi}~\cite{gurobi} (version 12.0.0).
We compare our approach against a standard mean squared error loss (\texttt{MSE}) and the DFL algorithms \texttt{SPO+}~\cite{elmachtoub_smart_2021,mandi2020smart} and \texttt{IMLE}~\cite{niepert_implicit_2021}. For these baselines, we rely on the implementations provided in \texttt{PyEPO}~\cite{tang2023pyepopytorchbasedendtoendpredictthenoptimize}. Predictive models are implemented and trained using \texttt{PyTorch}~\cite{paszke2019pytorchimperativestylehighperformance}. We also compare our approach with other methods focussed on scaling: \texttt{CaVE} and \texttt{CaVE+}~\cite{CaVE2024} as well as \texttt{LAVA}~\cite{berden2025solverfreedecisionfocusedlearninglinear}. These methods are less generic than ours, and cannot be applied to quadratic portfolio optimization. We do not compare with \texttt{E-PLL}~\cite{defresne2023scalablecouplingdeeplearning}, a solver-free neuro-symbolic approach for discrete graphical models (cost function networks), as it does not directly apply to our continuous quadratic portfolio nor to the global linear capacity constraints of the multi-dimensional knapsack; a detailed comparison on suitable discrete benchmarks is left for future work.
Our Lagrangian decomposition framework supports two modes, single and multiple decomposition (see the corresponding algorithms), with two loss functions available for each mode.  In addition, it can be integrated into any DFL method. 
Finally, because our approach introduces additional overhead for optimizing the multipliers when constructing the training set, we consider two training-budget settings. In the first, the total budget (including dataset construction) is matched to that of the corresponding DFL baseline ($=$). In the second, we allow the multipliers to be trained until convergence ($\infty$).
We denote our approaches as  $\texttt{LD}(\eta,\gamma,\mathcal{L},B)$, where $\eta \in \{\texttt{SPO+}, \texttt{IMLE} \}$, $\gamma \in \{\texttt{static}, \texttt{multiple}, \texttt{exact} \}$, $\mathcal{L} \in \{\mathcal{L}_1, \mathcal{L}_2\}$, and $B \in\{=,\infty\}$. Method \texttt{exact} refers to the exact differentiation we designed for the portfolio problem.  As evaluation metrics, we report the regret achieved by each approach over the test dataset, as commonly done in related works. Each model is trained with ten random seeds to assess the stability of the training procedure. The implementation is available at \url{https://github.com/corail-research/DFL-LD}.

\subsection{Results: Performances with Unlimited Budget}
This first experiment compares our approaches with prior baselines under an unrestricted training budget (denoted by the $\infty$ variant), meaning that all methods are trained to convergence, which is a commonly adopted assumption in the field. The results, reported as averages with 95\% confidence intervals over $10$ random seeds, are summarized in Table~\ref{tab:bookmark-log}. In all settings, we use early stopping and retain the checkpoint achieving the lowest validation relative regret, the training time reported in Table~\ref{tab:bookmark-log} corresponds to the wall-clock time at which this best validation model is reached.
We note that for the quadratic portfolio problem, the multiple-decomposition strategy is less relevant because one of the possible choices for the main subproblem (keeping the linear budget constraint) leads to a weakly informative optimization layer. We therefore focus on the decomposition that retains the quadratic risk structure (the subproblem forms are provided in Appendix~\ref{app:portfolio_subproblems}). Similarly, our $\mathcal{L}_2$ variant is less relevant for \texttt{SPO+}, as this differentiation technique is designed as a surrogate for regret and is therefore naturally aligned with the regret-like loss $\mathcal{L}_1$, whereas $\mathcal{L}_2$ does not correspond to a regret.
Overall, we observe that the best performance is consistently achieved by one of our proposed approaches. Interestingly, neither \texttt{IMLE} nor \texttt{SPO+} dominates across all settings, which is consistent with the findings of Mandi et al.~\cite{mandi_decision-focused_2024}. We view this as an important strength of our framework, as it can be combined with the most suitable decision-focused learning technique depending on the specific problem. Moreover, we observe that baselines designed to achieve low training times (\texttt{CaVE}, \texttt{CaVE+} and \texttt{LAVA}) perform worse than other DFL algorithms \texttt{IMLE} or \texttt{SPO+}. This observation may be explained by the noisy nature of the datasets. Although \texttt{CaVE} and \texttt{CaVE+} converge significantly faster than our approaches, our methods converge to models with superior final performance.  Finally, the exact LD-based differentiation
achieves the best results for the quadratic portfolio, showing the versatility of LD in DFL.

\subsection{Analysis: Training with a Restricted Budget}

\begin{figure}[!ht]
     \centering
     \begin{subfigure}[b]{0.9\columnwidth}
         \centering
         \includegraphics[width=\columnwidth]{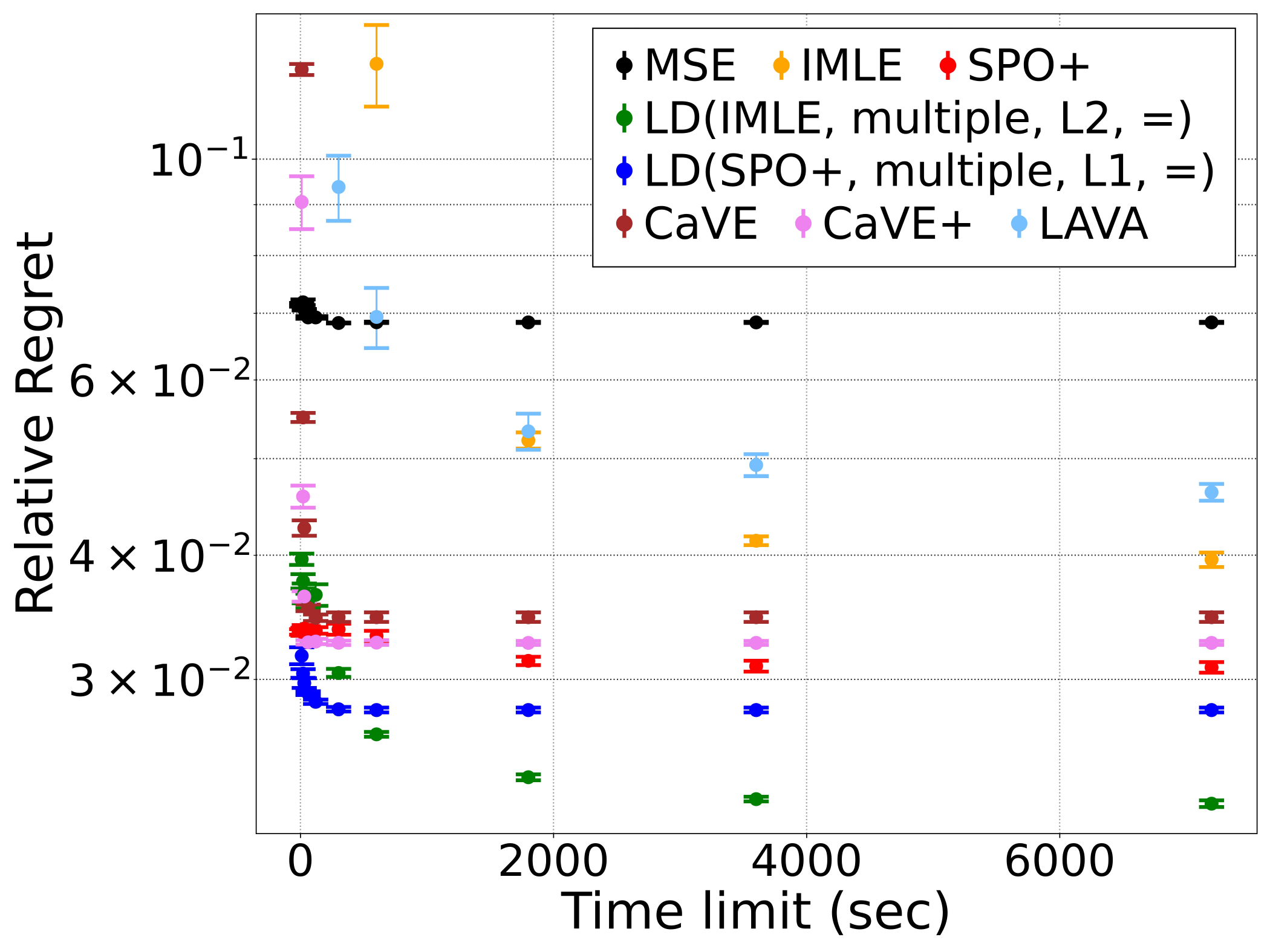}
         \caption{MD knapsack ($n=300$)}
         \label{fig:training-knapsack}
     \end{subfigure}
     \begin{subfigure}[b]{0.9\columnwidth}
         \centering
         \includegraphics[width=\columnwidth]{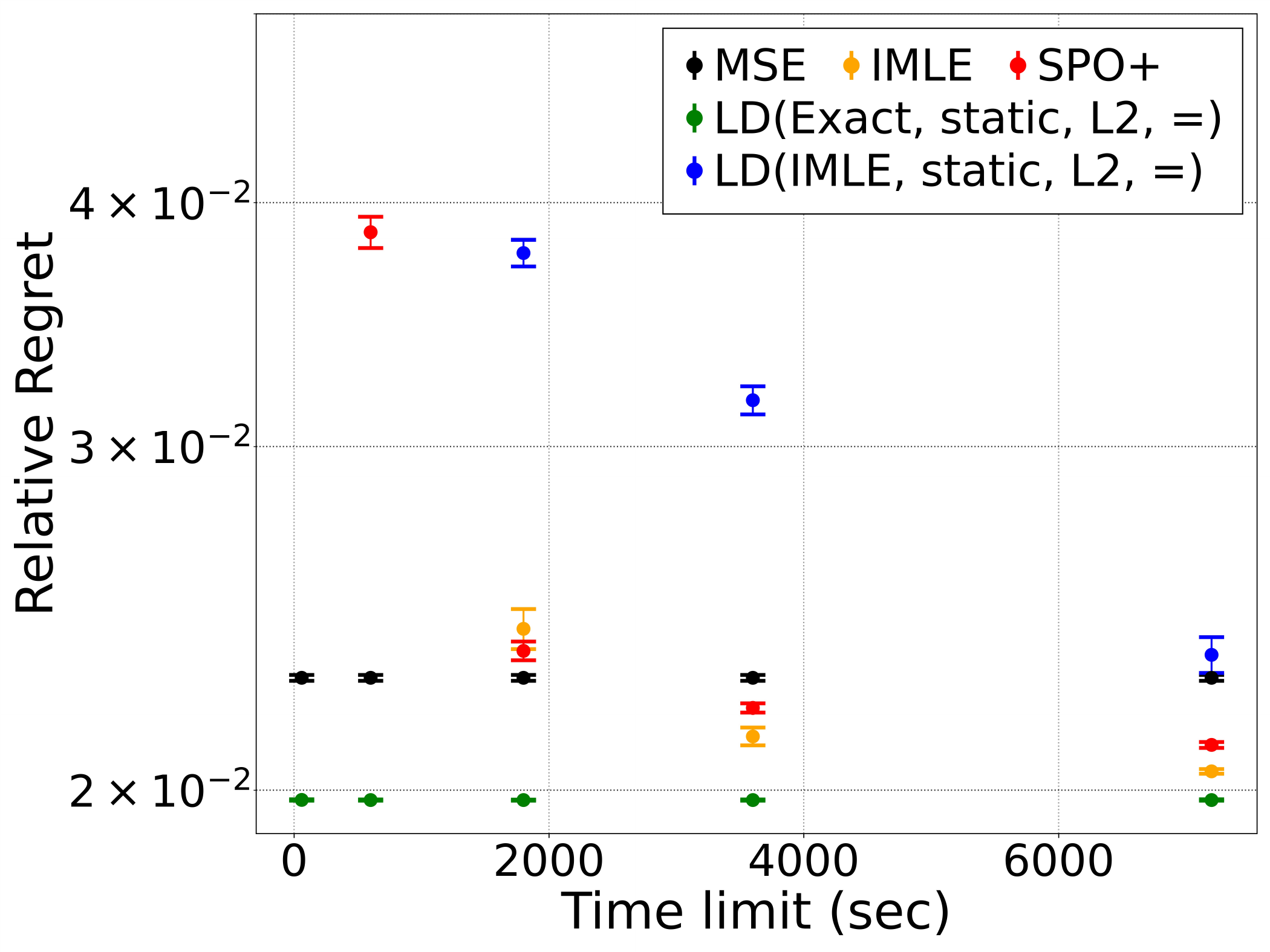}
         \caption{Portfolio ($n=400$)}
         \label{fig:training-portfolio}
     \end{subfigure}
     \caption{Test-set relative regret  (mean across seeds $\pm 95\%$ CI) with fixed training-time budget.}
\end{figure}

\begin{figure}[!ht]
     \centering
     \begin{subfigure}[b]{0.9\columnwidth}
         \centering
         \includegraphics[width=\columnwidth]{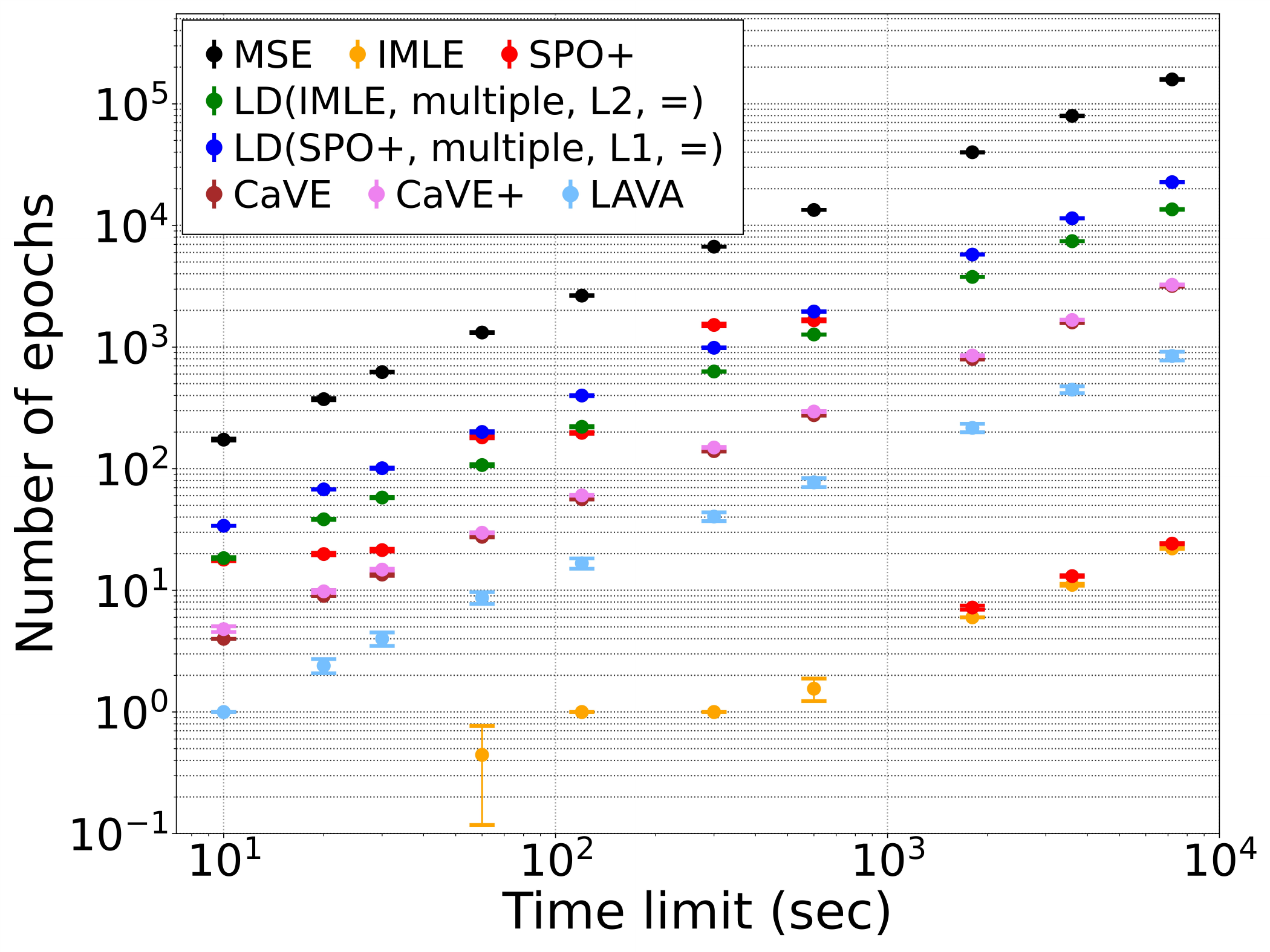}
         \caption{MD knapsack ($n=300$)}
         \label{fig:training-knapsack_epochs}
     \end{subfigure}

     \begin{subfigure}[b]{0.9\columnwidth}
         \centering
         \includegraphics[width=\columnwidth]{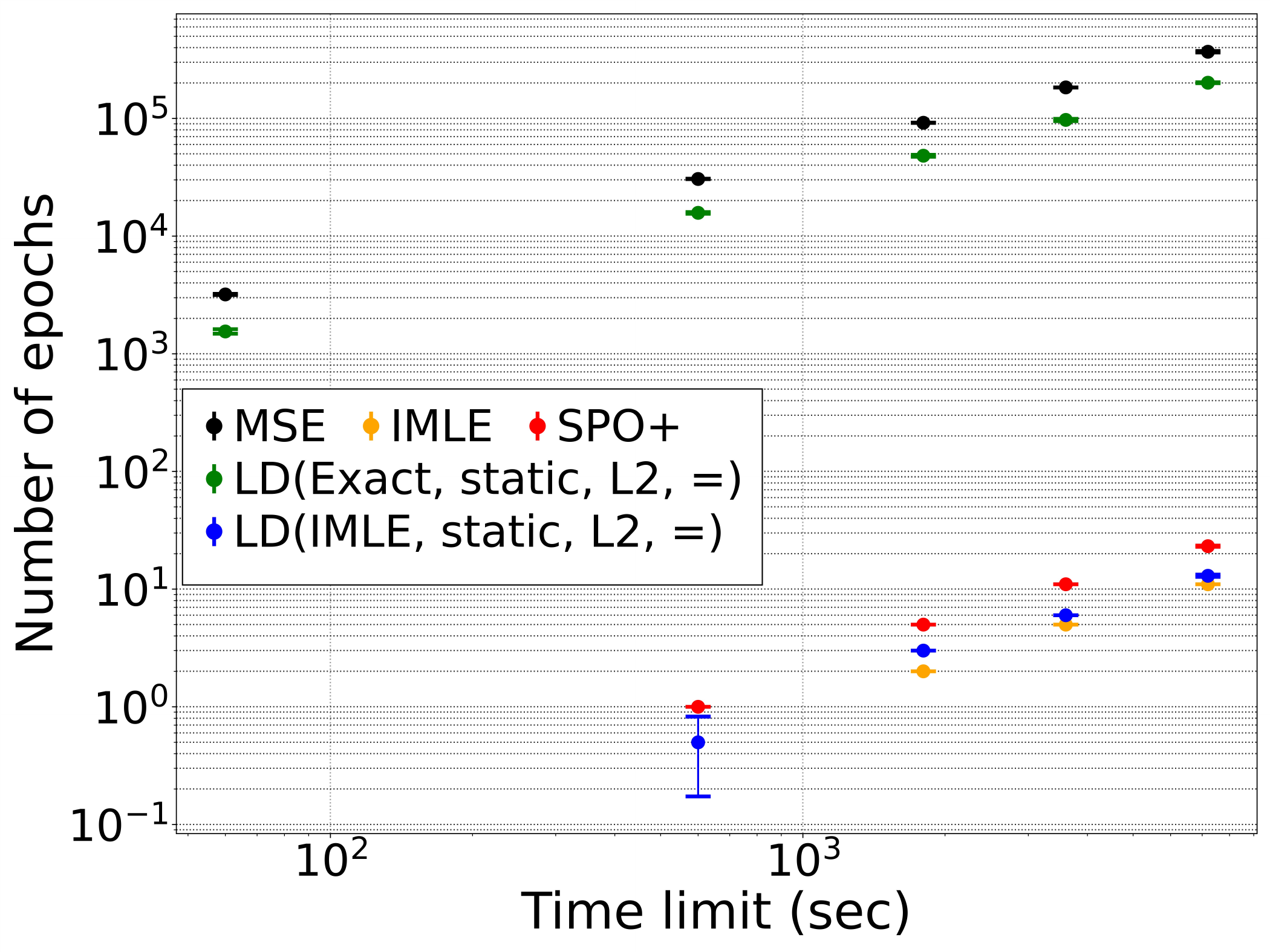}
         \caption{Portfolio ($n=400$)}
         \label{fig:training-portfolio_epochs}
     \end{subfigure}
     \caption{Number of epochs achieved during training  (mean across seeds $\pm 95\%$ CI) with fixed training-time budget.}
\end{figure}

This second set of experiments analyzes the behavior of the methods under a fixed training-time budget, where all approaches are allocated the same execution time. For our methods, this budget includes the construction of the training set (denoted by the $=$ variant). We restrict the analysis to the largest instances and to our best-performing approaches. The results are shown in Figures~\ref{fig:training-knapsack} and~\ref{fig:training-portfolio}. We report the relative regret on the test set for different training-time budgets. Hyperparameters are tuned independently for each budget (via validation), so each point corresponds to a different trained model rather than successive checkpoints of a single run. As expected, thanks to the Lagrangian decomposition, our approach solves the underlying optimization problems more efficiently at each training iteration, as the resulting subproblems are easier to solve. This allows training epochs to be executed faster, which ultimately leads to improved performance compared to the baselines under the same training-time budget. This results suggest that the Lagrangian multipliers optimization overhead can be controlled while remaining competitive. To confirm the scalability of our approach, we report the number of training epochs completed within each training-time budget in Figures~\ref{fig:training-knapsack_epochs} and~\ref{fig:training-portfolio_epochs}. The results directly corroborate this hypothesis, as at least one LD-based approach consistently achieves the highest number of epochs for a given time budget, among DFL baselines.

\subsection{Analysis:  Benefits of Parallelization}

Another benefit of the Lagrangian decomposition is that, in the multiple decomposition setting, each decomposition yields an independent sub-problem with its own set of Lagrangian multipliers. As a result, the offline dataset-construction step (i.e. computing these multipliers via subgradient iterations) is naturally amenable to parallelization. 
We evaluate the benefits of parallelization on the multi-dimensional knapsack problem with $200$ items. 
In all runs, we allocate 10 CPU cores, we then vary how many parallel worker processes are used to distribute the independent multiplier optimizations across these cores (from 1 worker, i.e., fully sequential, up to 10 workers, i.e., one per decomposition). Starting from an execution time of 4,218 seconds to perform 1,000 subgradient iterations per multiplier, we reduce the runtime to 882 seconds with 5 workers and to 462 seconds with 10 workers. 
This corresponds to a speed-up of approximately 9.12$\times$, indicating that the overhead induced by parallelization is negligible. 
We note that this parallelization opportunity is specific to the multiple-decomposition variant, where several independent multiplier optimizations must be carried out. In particular, it provides marginal benefit for the quadratic portfolio problem in which the multiple decomposition setting is not used.
Finally, these results show that our approach is well suited to scaling decision-focused learning to large-scale problems, provided they admit a Lagrangian decomposition, as is the case for the multi-dimensional knapsack and the quadratic portfolio problem.

\section{Conclusion}

Decision-focused learning is a promising paradigm for addressing predict-then-optimize problems. However, its practical deployment is often hindered by high computational costs, as it requires solving a constrained optimization problem for each training instance at every iteration. This significantly limits its ability to scale to large combinatorial problems. In this paper, we introduced a new methodology based on Lagrangian decomposition to substantially improve the scalability of DFL algorithms. We proposed a new surrogate objective and two associated loss functions for training the prediction model. Experiments on large-scale instances, up to eight times larger than those commonly considered in the literature, demonstrate that our approach can be seamlessly integrated into standard DFL methods, leading to improved performance. Scalability can be further enhanced through parallelization.
As future work, we plan to extend our approach to other combinatorial optimization problems in more realistic settings, such as shortest path and shift scheduling. Another promising direction is to investigate alternative strategies for updating the Lagrangian multipliers during training. 
Finally, establishing a theoretical relationship between the regret on the decomposed problem and the accuracy on the original primal problem is also an interesting direction to pursue.

\appendix

\section{Portfolio: Forms of the Main Subproblem}
\label{app:portfolio_subproblems}

The quadratic portfolio problem involves two constraints: the linear budget constraint
$\mathbf{1}^\top \mathbf{x}=1$ and the quadratic risk constraint
$\mathbf{x}^\top \boldsymbol{\Sigma}\mathbf{x}\le \gamma \bar{\Sigma}$.
In our Lagrangian-decomposition framework, we may choose either constraint to define the main subproblem. We detail the two resulting cases below.

\paragraph{Linear budget constraint as main constraint.}
If the budget constraint is selected, the main subproblem reads
\begin{equation}
\max_{\mathbf{X}_1 \ge 0}\; (\hat{\mathbf{c}}+\boldsymbol{\mu_2})^\top \mathbf{X}_1
\quad\text{s.t.}\quad \mathbf{1}^\top \mathbf{X}_1 = 1.
\end{equation}
This subproblem admits a closed-form solution: it places all the mass on an asset attaining
$\arg\max_i(\hat{c}_i+\mu_{2,i})$ (and zero elsewhere). As a result, it is computationally trivial and
does not capture the quadratic risk structure of the original formulation, which makes it less
informative for training.

\paragraph{Quadratic risk constraint as main constraint.}
If instead the quadratic constraint is selected, the main subproblem becomes
\begin{equation}
\max_{\mathbf{X}_1 \ge 0}\; (\hat{\mathbf{c}}+\boldsymbol{\mu_2})^\top \mathbf{X}_1
\quad\text{s.t.}\quad \mathbf{X}_1^\top \boldsymbol{\Sigma}\mathbf{X}_1 \le \gamma \bar{\Sigma}.
\end{equation}
Unlike the previous case, this formulation retains the covariance matrix $\boldsymbol{\Sigma}$ in
the optimization layer, and is therefore more faithful to the structure of the original quadratic
portfolio problem.

\section{The \textit{Exact} Method}
\label{app:Exact}

In the Lagrangian Decomposition, we choose the quadratic constraint in the main sub-problem. Thus, this sub-problem\footnote{Since there are only two constraints in the Portfolio problem, we have $\mu = (\boldsymbol{\mu_2})$.} is 
\begin{equation}
\label{eq:portfolio_main2}
\max_{\mathbf{X}_{1}\ge 0}
      \bigl(\mathbf{c}+\boldsymbol{\mu_2}\bigr)^{\top}\mathbf{X}_{1}
\quad
\text{s.t.}\quad
\mathbf{X}_{1}^{\top}\Sigma\mathbf{X}_{1}\le\gamma\bar{\Sigma}
\end{equation}

\paragraph{\textbf{Analysis of another problem.}}We consider the problem  
\begin{equation}
\label{eq:problem_exact}
\max_{\mathbf{w}\in \mathbb{R}^n}
      \mathbf{c}^{\top}\mathbf{w}
\quad
\text{s.t.}\quad
\mathbf{w}^{\top}\Sigma\mathbf{w}-\gamma\bar{\Sigma} \leq 0
\end{equation}

We define $J(\mathbf{w})=\mathbf{c}^{\top}\mathbf{w}$, $F(\mathbf{w}) = \mathbf{w}^{\top}\Sigma\mathbf{w}-\gamma\bar{\Sigma}$ and $K$ the space of feasible solutions.

We note that the $J$ and $F$ are continuous and convex with respect to $\mathbf{w}$. Moreover, $F'(\mathbf{w})$ is nonzero for all $\mathbf{w} \in K$ with $\mathbf{w} \ne \mathbf{0}$, which ensures that the constraint qualification condition is satisfied.

We can thus apply the Kuhn and Tucker Theorem. For any nonzero $\mathbf{w} \in K$, $\mathbf{w}$ is a global maximizer of $J$ over $K$ if and only if 
\begin{align}
\label{eq:kkt}
\exists p \geq 0 \,, F(\mathbf{w}) \leq 0  \,, p \cdot F(\mathbf{w}) = 0 \,,\mathbf{J}'(\mathbf{w}) - p \mathbf{F}'(\mathbf{w}) = \mathbf{0} \,
\end{align}

We have 
\begin{align*}
\eqref{eq:kkt} \;\Longleftrightarrow\; &
\exists p \geq 0 \,,\quad F(\mathbf{w}) \leq 0,\quad
p \cdot F(\mathbf{w}) = 0,\\
&\mathbf{c} - 2p\Sigma\mathbf{w} = \mathbf{0}
\end{align*}

Since $\mathbf{c} \ne \mathbf{0}$, we must have $p \ne 0$. It follows that
\begin{align*}
\eqref{eq:kkt} \;\Longleftrightarrow\;&
\exists p \geq 0 \,,\quad F(\mathbf{w}) \leq 0,\quad
F(\mathbf{w}) = 0,\\
&\mathbf{c} - 2p\,\Sigma\mathbf{w} = \mathbf{0}\\
\Longleftrightarrow\;&
\exists p \geq 0 \,,\quad F(\mathbf{w}) \leq 0,\quad
\mathbf{w}^{\top} \Sigma \mathbf{w} = \gamma \, \bar{\Sigma},\\
&\mathbf{w} = \frac{1}{2p} \, \Sigma^{-1} \mathbf{c}\\
\Longleftrightarrow\;&
\exists p \geq 0 \,,  F(\mathbf{w}) \leq 0,
\frac{1}{4p^2} \left( \Sigma^{-1} \mathbf{c} \right)^{\top} \Sigma \Sigma^{-1} \mathbf{c} = \gamma \bar{\Sigma},\\
&\mathbf{w} = \frac{1}{2p} \, \Sigma^{-1} \mathbf{c}\\
\Longleftrightarrow\;&
\exists p \geq 0 \,,\quad F(\mathbf{w}) \leq 0,\quad
p = \frac{1}{2} \sqrt{ \frac{ \mathbf{c}^\top \Sigma^{-1} \mathbf{c} }{ \gamma \bar{\Sigma} } },\\
&\mathbf{w} = \sqrt{ \frac{\gamma \bar{\Sigma}}{\mathbf{c}^\top \Sigma^{-1} \mathbf{c}} }\, \Sigma^{-1} \mathbf{c}
\end{align*}

We conclude that the optimal solution of \eqref{eq:problem_exact} is $\mathbf{w}^* = \sqrt{ \frac{\gamma \bar{\Sigma}}{\mathbf{c}^\top \Sigma^{-1} \mathbf{c}} }\, \Sigma^{-1} \mathbf{c}$.\\

\paragraph{\textbf{Back to our sub-problem.}} Once the constraint $\mathbf{X}_{1}\geq \mathbf{0}$ is removed, the solution to
\begin{equation}
\label{eq:relax_main}
\max_{\mathbf{\bar{X}}_{1}\in\mathbb{R}^n}
      \bigl(\mathbf{c}+\boldsymbol{\mu_2}\bigr)^{\top}\mathbf{\bar{X}}_{1}
\quad
\text{s.t.}\quad
\mathbf{\bar{X}}_{1}^{\top}\Sigma\mathbf{\bar{X}}_{1}\le\gamma\bar{\Sigma}
\end{equation}

is given by 
\begin{equation}
\mathbf{\bar{X}_1}^*(\mathbf{c}, \mu) = \sqrt{ \frac{\gamma \bar{\Sigma}}{\bigl(\mathbf{c}+\boldsymbol{\mu_2}\bigr)^\top \Sigma^{-1} \bigl(\mathbf{c}+\boldsymbol{\mu_2}\bigr)} }\, \Sigma^{-1} \bigl(\mathbf{c}+\boldsymbol{\mu_2}\bigr)
\end{equation}

We observe that the closed-form solution is differentiable with respect to $\mathbf{c}$, and its gradient can be obtained directly via PyTorch’s automatic differentiation. These findings motivate a problem-specific differentiation scheme for the portfolio model. During training we replace the solver-provided solution $\mathbf{X}_1^{*}(\mathbf{c},\mu)$ with the analytical expression $\mathbf{\bar{X}}_1^{*}(\mathbf{c},\mu)$.  This eliminates the need for a solver and a generic DFL technique.

\section*{Acknowledgments} This research has been supported by the CRM and the Simons Foundation, thematic program Combinatorial Optimization and Data Science.
It has been partly funded by the European Research Council (ERC) under the EU Horizon 2020 research and innovation program (Grant No 101002802, CHAT-Opt)
and by the Canada Research Chairs Program in Healthcare Analytics. We also acknowledge the support of the Natural Sciences and Engineering Research Council of Canada (NSERC RGPIN-2025-04995 and RGPIN-2022-03964).

\bibliographystyle{named}
\bibliography{dfl-bibli}


\end{document}